\def\cvprPaperID{64}
\def\confName{ICCV}
\def\confYear{2023}

\def\paperTitle{Accelerating Deep Neural Networks via Semi-Structured Activation Sparsity}

\def\authorBlock{
    Matteo Grimaldi \qquad 
    Darshan C. Ganji \qquad 
    Ivan Lazarevich \qquad 
    Sudhakar Sah \\
    Deeplite \\
    {\tt\small matteo.grimaldi@deeplite.ai}
}

\newif\ifreview 
\newif\ifarxiv \newcommand{\arxiv}{\arxivtrue}
\newif\ifcamera 
\newif\ifrebuttal 

\arxiv

\pdfoutput=1
\documentclass[10pt,twocolumn,letterpaper]{article}
\ifreview \usepackage[review]{cvpr} \fi
\ifarxiv \usepackage[pagenumbers]{cvpr} \fi
\ifrebuttal \usepackage[rebuttal]{cvpr} \fi
\ifcamera \usepackage{cvpr} \fi

\usepackage{graphicx}
\usepackage{amsmath}
\usepackage{amssymb}
\usepackage{booktabs}

\usepackage{times}
\usepackage{microtype}
\usepackage{epsfig}
\usepackage[table,xcdraw]{xcolor}
\usepackage{caption}
\usepackage{float}
\usepackage{placeins}
\usepackage{color, colortbl}
\usepackage{stfloats}
\usepackage{enumitem}
\usepackage{tabularx}
\usepackage{xstring}
\usepackage{multirow}
\usepackage{xspace}
\usepackage{url}
\usepackage{subcaption}
\usepackage{xcolor}
\usepackage[hang,flushmargin]{footmisc}
\usepackage[linesnumbered,algoruled]{algorithm2e}
\usepackage{comment}
\usepackage{placeins}
\usepackage{booktabs}
\usepackage{multirow}

\ifcamera \usepackage[accsupp]{axessibility} \fi

% ...

\ifarxiv  \fi

\newcommand{\R}[1]{{%
    \textbf{%
        \ifstrequal{#1}{1}{\textcolor{red}{R#1}}{%
        \ifstrequal{#1}{2}{\textcolor{blue}{R#1}}{%
        \ifstrequal{#1}{3}{\textcolor{magenta}{R#1}}{%
        \ifstrequal{#1}{4}{\textcolor{teal}{R#1}}{%
                           \textcolor{cyan}{R#1}%
        }}}}%
    }%
}}

\usepackage{xr-hyper}

\makeatletter
\newcommand*{\addFileDependency}[1]{
  \typeout{(#1)}
  \@addtofilelist{#1}
  \IfFileExists{#1}{}{\typeout{No file #1.}}
}

\makeatother

\usepackage[pagebackref,breaklinks,colorlinks]{hyperref}
\usepackage[capitalize]{cleveref}
\crefname{section}{Sec.}{Secs.}
\crefname{table}{Table}{Tables}
\crefname{figure}{Fig.}{Figs.}

\begin{document}
\title{\paperTitle}
\author{\authorBlock}
\maketitle
\begin{abstract}

The demand for efficient processing of deep neural networks (DNNs) on embedded devices is a significant challenge limiting their deployment. Exploiting sparsity in the network's feature maps is one of the ways to reduce its inference latency. It is known that unstructured sparsity results in lower accuracy degradation with respect to structured sparsity but the former needs extensive inference engine changes to get latency benefits.
To tackle this challenge, we propose a solution to induce semi-structured activation sparsity exploitable through minor runtime modifications. To attain high speedup levels at inference time, we design a sparse training procedure with awareness of the final position of the activations while computing the General Matrix Multiplication (GEMM). We extensively evaluate the proposed solution across various models for image classification and object detection tasks. Remarkably, our approach yields a speed improvement of $1.25 \times$ with a minimal accuracy drop of $1.1\%$ for the ResNet18 model on the ImageNet dataset. Furthermore, when combined with a state-of-the-art structured pruning method, the resulting models provide a good latency-accuracy trade-off, outperforming models that solely employ structured pruning techniques. The code is available at~{\small\url{https://github.com/Deeplite/activ-sparse}} .

\end{abstract}
\section{Introduction}
\label{sec:intro}

Deep neural networks (DNNs) have become the go-to state-of-the-art solution in most domains of machine learning in recent years, like computer vision~\cite{krizhevsky2012imagenet}, natural language understanding~\cite{vaswani2017attention} and generative AI~\cite{koubaa2023gpt}. Oftentimes, the computational footprint of DNN models limits their usage on low-resource embedded processors. Compression and acceleration of such models is an active research area aimed at bridging this gap~\cite{survey2020modelcompression} and could be generally categorized into pruning ~\cite{cnn_network_slimming, prune_or_not_to_prune, pruning_efficient_convnet}, tensor decomposition \cite{tucker_decomposition_liu2022deep}, quantization~\cite{esser2019learned, quantiztion_whitepaper}, development of lightweight neural networks~\cite{wide_reduced_prec_nw, mobilenet, squeezenet}, and runtime optimizations~\cite{tvm, efficient_inf_engine}. 

Pruning remains a prominent compression method, particularly evidenced by recent strides in structured weight pruning, achieving  state-of-the-art latency-accuracy trade-offs across diverse computer vision tasks~\cite{depgraph}. However, existing research in pruning has predominantly focused on removing redundant model parameters, overlooking the potential inherent sparsity within feature maps, commonly referred to as activations.
Activation sparsity is naturally intrinsic in DNNs with ReLU-like activation functions to a certain extent~\cite{rhu2018compressing, li2022lazy}. 
Nevertheless, this sparsity, tied to the functional form of the ReLU non-linearity, retains an unstructured nature and lacks homogeneity across layers. 
Several methods have emerged to artificially augment activation sparsity during training, enhancing model generalization and robustness through regularization techniques~\cite{dropblock, revisiting_dropout}. However, such methods selectively remove blocks of connected pixels solely during model training, maintaining denseness at inference time and consequently forfeiting opportunities for model inference acceleration. In contrast, to achieve faster model execution post-training, activation sparsity needs to extend to inference time as well.
A variety of works explored {\em data-dependent} mechanisms to exploit activation sparsity at runtime, dynamically selecting the pixels according to the complexity of the input sample to process~\cite{adaptive_pixelwise_activation_sparse, ren2018sbnet, more_is_less_gemm_dong2017more}. While these approaches efficiently reduce computations with minimal accuracy loss, effectively integrating them into low-power embedded devices can be challenging due to the required architectural modifications.
In contrast, {\em data-free} strategies employ custom regularization with proper hard-thresholding to establish a fixed and constant sparsity pattern~\cite{activation_map_compression_georgiadis2019, inducing_activation_sparsity}. Such a strategy guarantees consistent speedup across distinct input samples. However, the absence of structured regular patterns among zeroed elements confines these model acceleration benefits to dedicated sparse inference engines (e.g., DeepSparse~\cite{inducing_activation_sparsity}).

To tackle these challenges, we propose an efficient DNN compression pipeline that consists of (i) a novel training scheme that induces semi-structured sparsity in activation feature maps and (ii) an easy-to-implement runtime modification that allows exploiting the semi-structured sparsity of the network's activations at inference time. The proposed sparsity pattern for feature maps is structured in the channel dimension, but unstructured in the spatial dimension. That is, a set of individual pixels are zeroed across all channels of the feature map. We suggest an effective way to construct such sparsity masks during training and demonstrate how these sparse masks can be used by the runtime during inference. With XNNPACK~\cite{xnnpack} as an example library, we implement a runtime modification that transforms the semi-structured sparsity of activations into effectively structured sparsity, resulting in reduced computational load through the use of lower ranks in General Matrix Multiplication (GEMM). 

To summarize, the primary focus of this study could be outlined as follows:

\begin{itemize}
    \item We propose a novel training scheme inducing semi-structured activation sparsity in deep neural networks via the propagation of random spatial masks.
    \item We show that sampling of random masks during training followed by mask freezing improves the performance of DNNs under the constraint of semi-structured sparsity in activations.
    \item We demonstrate the effectiveness of the proposed training scheme on image classification and object detection tasks and show how it can be combined with structured pruning to get a competitive accuracy-latency trade-off.
    \item We provide an example of an easy-to-implement runtime modification on top of XNNPACK \cite{xnnpack} that allows obtaining latency speedup of up to 2$\times$ with relatively low sparsity rates (under $50\%$).
\end{itemize}

\section{Related Work}
\label{sec:related}

Over the past few years, significant progress has been made in the field of deep learning model compression and acceleration, aimed at improving the efficiency of deep neural networks during inference by reducing their memory and computational requirements. 
Pruning~\cite{prune_or_not_to_prune, cnn_network_slimming} focuses on removing redundant connections or units in the model architecture based on heuristic importance criteria, resulting in streamlined models with improved efficiency.
Quantization~\cite{wide_reduced_prec_nw, jacob2018quantization} tackles model size compression by reducing the numerical precision of weights and activations from standard $32$-bit floating-point representations to lower bit-widths such as $8$-bit, or in more extreme cases, $2$-bit or $1$-bit.
Knowledge distillation~\cite{distilling_hinton_2015, distillation_zeng2000using} involves transferring knowledge from a larger, more complex network to a smaller one, allowing the compact model to attain comparable performance to its larger counterpart.
Hand-crafted models, exemplified by architectures like MobileNetV3~\cite{howard2019searching}, EfficientNetV2~\cite{tan2021efficientnetv2} and ShuffleNetV2~\cite{ma2018shufflenet}, are often designed with custom operations and blocks optimized for faster inference, thereby enhancing overall efficiency.
Furthermore, apart from direct model modifications, there are other strategies aimed at improving the efficiency of deep neural networks.
Graph order rewriting involves transforming the network's computational graph to optimize its execution flow, thus enhancing overall performance~\cite{ahn2020ordering}. 
Custom runtime optimization~\cite{tvm, efficient_inf_engine} aims to maximize model performance at the operator level, harnessing the target hardware's potential. It becomes indispensable in cases where existing operators or processing units cannot directly execute certain model structures, such as unstructured sparse or low-bit quantized models, requiring specific adaptations for seamless and efficient execution.

\subsection{Pruning}
Pruning methods can be usually categorized according to their granularity~\cite{hoefler2021sparsity} or to their importance policy.
In terms of granularity, pruning can usually operate with {\em unstructured} or {\em structured} sparsity patterns. Unstructured pruning involves removing single connections in the network based on their importance~\cite{han2015deep, molchanov2017variational}. Targeting individual weights offers flexibility in achieving high accuracy but may lead to challenges in efficient inference due to irregular memory access patterns. A custom runtime with specialized sparse kernels is often necessary to achieve speedup in case of unstructured sparsity (e.g., DeepSparse~\cite{deepsparse}). Conversely, structured pruning~\cite{molchanov2016pruning, pruning_efficient_convnet_li2016} involves the removal of entire channels or filters from the network, which can pose challenges during model training due to its more substantial impact on accuracy. However, pruning at this level of granularity can significantly enhance model efficiency in many existing runtimes, resulting in notable reductions in storage requirements and accelerated inference latency.

Pruning policies encompass various schemes and criteria for efficient model compression.
Magnitude-based criteria rely on the absolute weight values to identify less important parameters~\cite{han2015deep, filter_level_pruning}, while first-order methods leverage gradients for importance ranking~\cite{dong2017learning, molchanov2019importance}. Some approaches involve one-time pruning followed by retraining~\cite{han2015learning}, while others adopt iterative pruning techniques~\cite{pruning_efficient_convnet}. 
Recent research has explored the efficacy of various pruning methods, offering valuable insights to enhance model compression techniques~\cite{pruing_confusing_wang2023state}. Notably, DepGraph~\cite{depgraph} introduced a novel method for general structural pruning of arbitrary architectures, efficiently removing coupled parameters for model acceleration. The results demonstrate its superior performance compared to many other techniques.

\begin{figure*}[!ht]
    \centering
    \includegraphics[width=0.99\textwidth]{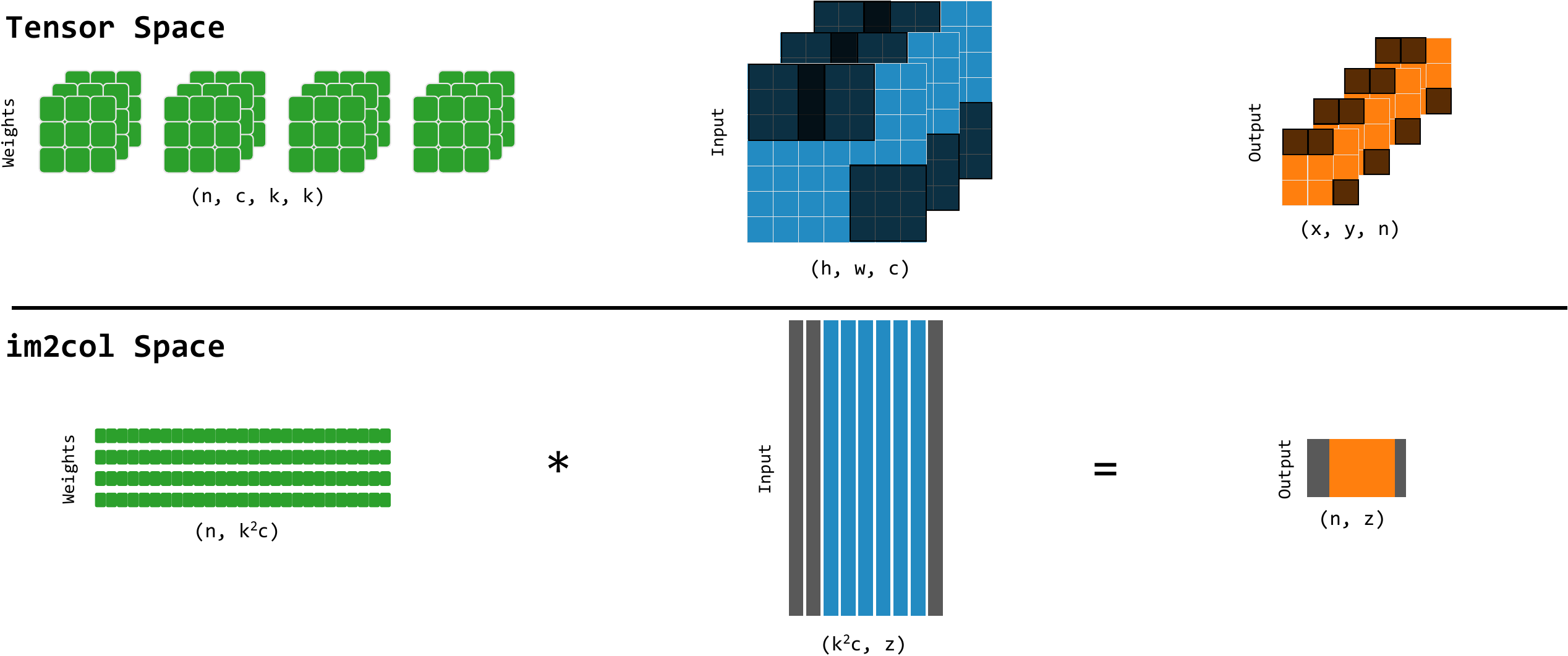}
    \caption{Illustration of the proposed activation sparsity pattern in both tensor and \texttt{im2col} spaces.}
    \label{fig:conv_output_masking}
\end{figure*}

\subsection{Activation Sparsity}
Another crucial sphere of inquiry revolves around exploiting the inherent sparsity present within neural network feature maps, particularly in the context of computer vision applications. The induction of activation sparsity stands out as a pivotal technique for latency reduction, providing a synergistic complement to weight pruning strategies. Sparsity is naturally present in feature maps due to the presence of ReLU-like activation functions which force feature maps to become zero when their values fall below certain thresholds~\cite{inducing_activation_sparsity,li2022lazy}. 

The majority of efforts in the literature have been directed towards harnessing activation sparsity through {\em data-dependent} mechanisms, tightly linked to input complexity. This strategy entails an informed masking approach, where the sparsity pattern is dynamically generated based on the distribution of less informative pixels within the input samples. Consequently, a distinct sparsity pattern is generated for each input. Some of these techniques necessitate architectural adjustments for on-the-fly pattern generation at runtime\cite{adaptive_pixelwise_activation_sparse, ren2018sbnet, more_is_less_gemm_dong2017more}. Unfortunately, these requirements significantly hamper their effectiveness when deployed on resource-constrained devices.
As a result of these constraints, many of these works often lack real-world hardware validation or predominantly demonstrate latency improvements on higher-performance hardware configurations. For instance, the efficacy of sparsity has been pronounced in GPU deployment scenarios, yielding impressive latency enhancements such as up to $1.88\times$ acceleration on a ResNet50 architecture using a {\em Mali} GPU~\cite{activation_sparsity_gpu}. Similarly, the work by Xu et. al~\cite{act_sparse_conv_gpu_xu2019g} tailored custom kernels for Nvidia GPUs, resulting in performance acceleration of $3$-$4\times$.

In more recent investigations, novel regularization strategies have emerged to induce activation sparsity featuring a regular and consistent pattern, regardless of varying input samples ({\em data-free} strategies).
Georgiadis et. al~\cite{activation_map_compression_georgiadis2019} proposed to combine sparsity, quantization, and entropy encoding of activation maps to achieve up to $1.6\times$ inference acceleration and up to $6\times$ reduction of the memory footprint for architectures like InceptionV3 and MobileNetV1. Kurtz et al.~\cite{inducing_activation_sparsity} introduced a new regularization technique and threshold-based sparsification based on a parameterized activation function to maximize sparsity with minimal accuracy drop.
While these works are the most similar to our approach, they predominantly emphasize unstructured sparsity among zeroed elements. As a consequence, these model acceleration benefits remain confined to dedicated sparse inference engines like DeepSparse~\cite{inducing_activation_sparsity}.

\subsection{Low-Rank GEMM}
The widely adopted \texttt{im2col}-based General Matrix Multiply (GEMM) technique converts feature maps into column-wise matrices.
This transformation paves the way for streamlined matrix multiplication with weight matrices, thus fostering parallel computations and refining the convolutional operations.
Moreover, the low-rank GEMM approach focuses on reducing the number of rows (or columns) in one of the two matrices, aiming to decrease computational complexity and memory demands. 
Dong et al.~\cite{more_is_less_gemm_dong2017more} devised a trainable module learning collaborative kernels to selectively skip activation pixels during computation, yielding a $1.2\times$ speedup. Their analysis focused on two models and relatively simple datasets. 
In the context of video processing, the Skip-conv network~\cite{skip_conv} leverages residual images, creating sparsity exploited by low-rank GEMM. This approach suits moving objects, producing notable sparsity.
Liu et al.~\cite{adaptvie_sparse_inference_liu2020deep} applied sparse adaptive inference for super-resolution, more similar to our approach, but just tailored to low-rank GEMM for specific patches crucial in super-resolution tasks.

\section{Methodology}
\label{sec:method}

GEMM-based implementation of the convolution operation is typically favored over the direct one as GEMM enables faster and more efficient matrix operations, making it a preferred choice for deep learning inference engines.
Reducing the rank of the matrices in GEMM operations is generally directly correlated with faster computation, especially on low-power CPUs. Our proposed technique aims to reduce the rank of the input activation matrix (activation feature map in the \texttt{im2col} space) to speed up model inference. This is pursued by inducing semi-structured sparsity in the network at training time which will be exploited through lower-rank GEMMs at inference time.

Figure~\ref{fig:conv_output_masking} shows the convolution-as-GEMM implementation for convolutional layers, where both weights (green) and activations (blue) are unfolded respectively from $4$-D and $3$-D tensors to $2$-D matrices. The picture shows the standard convolution operation both in the tensor space (i.e., the standard space before the reshaping) and in the \texttt{im2col} space.
Each of the $n$ filters is reshaped into a row of $k^2c$ size, where $k$ is the kernel size and $c$ is the number of channels. In the same way, the input feature map is reshaped into a $k^2c \times z$ matrix, where each column is composed of all the pixels of the input sliding window ($k^2c$). The number of rows $z$ depends on the convolution parameters (e.g., stride, padding, and dilation values). Then a standard matrix multiplication of weights and activation matrices is computed to generate an $n \times z$ output  matrix. 

In order to reduce the rank of the activation matrix, a subset $s<z$ of columns needs to be removed. These columns correspond to elements covered by the sliding local tiles (covering all channels) used during the convolution in the tensor space.
To remove the columns at compute time, during each convolution, a subset $s$ of the sliding local tiles needs to be skipped: a binary mask with a \texttt{im2col}-based pattern is used to apply hard thresholding to the activation tensors, where the $s$ sparse columns of the activation matrix will be directly skipped during inference.
In the two following subsections, we show how to induce (at training time) and how to exploit (at inference time) such semi-structured activation sparsity.

\subsection{Training}
To induce activation sparsity with the \texttt{im2col} pattern, we need to group activations in the tensor space according to their final position after the \texttt{im2col} reshaping. We consider this approach as semi-structured as it is unstructured in the $width \times height$ space (spatial dimensions of the feature map) but it is structured across the channel dimension. 

\begin{figure}
    \centering
    \includegraphics[width=0.95\linewidth]{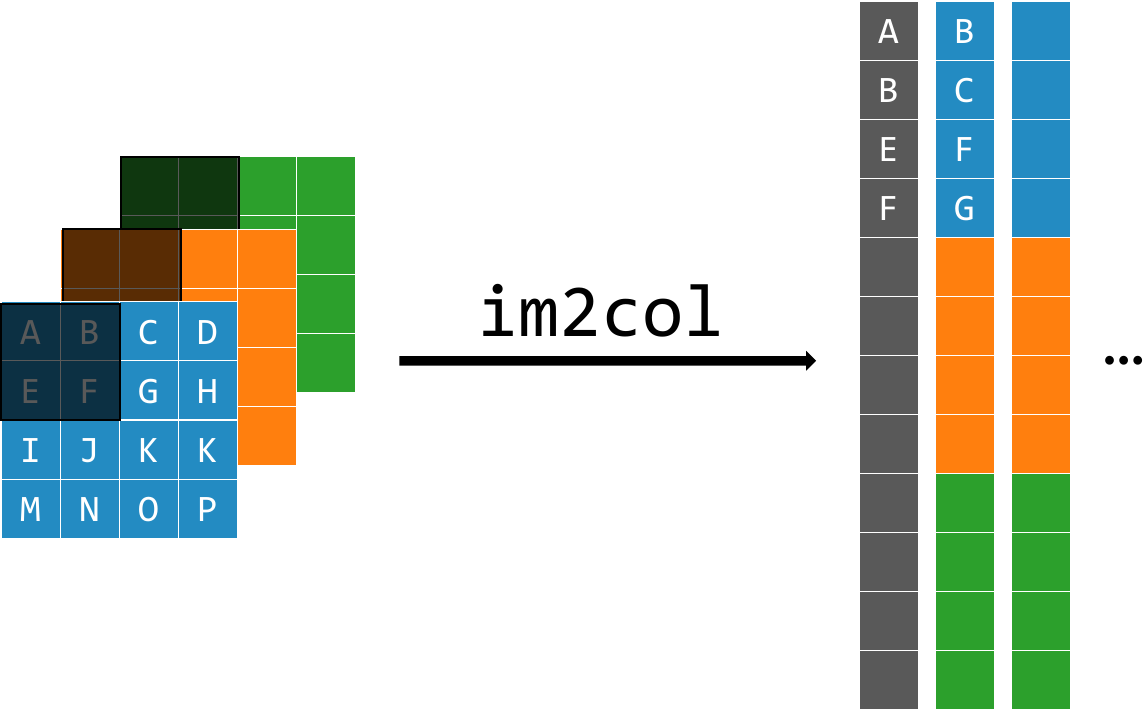}
    \caption{Example of the \texttt{im2col} procedure: input activations (left) and the activation matrix after transformation (right). Note that masking (highlighted in black) a sliding tile of the convolution affects only a single column in the reshaped matrix. In the first column, pixels $B$ and $F$ are masked, while they remain non-zero in the second column.}
    \label{fig:drop_pixel_conv_issues}
\end{figure}

Pruning activations with this pattern is a more delicate procedure compared to standard unstructured weight pruning, as the elements of the activation feature map cannot be directly removed from the model. The sparsified elements in the activations for one convolutional window/tile (i.e., one \texttt{im2col} column) could be kept dense (unmasked) for the next windows/tiles.
Figure~\ref{fig:drop_pixel_conv_issues} demonstrates this concept for a case when a single window (tile) is selected to be sparsified (masked). In this case, the pixels $\{A, B, C, D\}$ are dropped from the computation (including all the pixels/elements with the same $(width, height)$ coordinates in the other channels). This results in the first column of the \texttt{im2col} matrix becoming zero, which reduces the rank of the matrices to be multiplied. However, dropping (masking) this block from the feature map altogether should also affect the second column of the matrix, which is not selected to be pruned. For this reason, the pixels $B$ and $F$ will be masked for the first column but will be kept non-zero in the second one.

Introducing activation sparsity in deep neural networks for computer vision is challenging due to the varying positions of the regions of interest in images. Uniformly enforcing sparsity with a fixed pattern across data samples can lead to information loss for some images and retention for others. Achievable sparsity levels (while keeping accuracy degradation low) are often limited compared to weight sparsity, due to the dynamic and context-dependent nature of activation patterns in different input images.
It has been shown that inducing structured sparsity through sampling random masks~\cite{dropblock} can act as a regularizer that enhances the model's generalization and robustness. We found sampling random masks during training can reduce the accuracy loss when the sparsity rates are kept relatively low. The random ranking mechanism ensures that the selection of pixels to be masked is unbiased, contributing to the robustness of the training process. 
We propose a novel custom random masking approach, which involves randomly selecting a percentage of pixels from the input image to be masked. The resulting input image mask is then propagated consistently across all layers (employing pooling operations when downsampling is necessary). By propagating this initial random sparse pattern layer-to-layer, we ensure the preservation of the same masking structure throughout the network. This guarantees translation invariance across the feature maps of different layers, even when they have varying resolutions.
The proposed custom random mask sampling is a crucial aspect of our training procedure as it helps the model to prevent over-fitting to specific patterns and encourage more generalized learning, yet limiting accuracy loss. 
The generated binary masks, specific to each sparsity level, enable the model to adapt its weights during training, effectively promoting the benefits of sparsity while maintaining crucial representational capacity.
The training process comprises three key stages: (i) initially, a few dense pretraining epochs are performed; (ii) subsequently, our masking technique is applied gradually according to a schedule, incrementing sparsity rate until the desired target~\cite{prune_or_not_to_prune} is achieved; (iii) finally, the mask freezing stage ensues, where binary masks for each layer are fixed for the rest of the training process, allowing the model to recover from accuracy loss through more precise updates.

Algorithm~\ref{alg:training} outlines our sparse training pipeline. The algorithm takes the fixed sparsity percentage $s$ as an input and returns the trained model with a binary constant mask $mask$. 
The pruning scheduler (line $3$) controls the switch between dense (line $8$) and sparse forward steps (line $6$). 
The \texttt{updateMask} (line $4$) scheduler sets when to update or freeze the masks through the \texttt{getMask} function (line $5$). 
This mask is used by \texttt{maskedForward} to induce the sparsity in the feature maps.
At the end of the training, both the model and the masks are returned (line $11$). It needs to be highlighted that model weights are kept fully dense, and no weights are pruned.
The \texttt{getMask} function plays a critical role in our sparse training pipeline, as it is responsible for generating a different binary mask for each forward step.
At first, a random $2$-D score is generated according to the input image resolution (line $13$). This is propagated through the layers, downscaling the resolution when needed (lines $15$-$16$). At last, the function ranks the model's score and generates the binary mask (lines $17$-$19$). 
\setlength{\textfloatsep}{5pt}
\begin{algorithm}[t]
\SetKwFunction{pruneStep}{pruneStep}
\SetKwFunction{getRandomMask}{getRandomMask}
\SetKwFunction{getMask}{getMask}
\SetKwFunction{oneslike}{ones\_like}
\SetKwFunction{backward}{backward}
\SetKwFunction{forward}{forward}
\SetKwFunction{maskedForward}{maskedForward}
\SetKwFunction{maskedConv}{maskedConv}
\SetKwFunction{Conv}{Conv}
\SetKwFunction{FMain}{main}
\SetKwFunction{randomScore}{randomScore2d}
\SetKwFunction{avgpool}{avg\_pool2d}
\SetKwFunction{rankPixels}{rankPixels}
\SetKwFunction{oneslike}{ones\_like}
\SetKwFunction{updateMask}{updateMask}

  \SetKwProg{Fn}{Function}{:}{}
  \Fn{\FMain{model, steps, s}}{
    \For{t in steps} 
      { 
        \uIf {\pruneStep(t)} 
        {
            \uIf {\updateMask(t)}
            {
                mask = \getMask(model, s) \\
            }
            \maskedForward(model, mask) \\
        }
        \uElse
        {
            \forward(model) \\ 
        }
        \backward(model) \\
      }
      \KwRet model, mask \\
  }
  \BlankLine
  \SetKwProg{Pn}{Function}{:}{\KwRet }
  \Pn{\getMask{model, s }}{
    score = \randomScore(model.input\_res) \\
    \For{layer in model} 
      { 
        ratio = input\_res // layer.res \\
        layer\_score = \avgpool(score, ratio) \\
        idx = \rankPixels(layer\_score) \\
        mask = \oneslike(model) \\
        mask[idx] = $0$ \\
      }
        \KwRet mask \\
    }
\caption{Sparse Training}
\vspace{1mm}
\label{alg:training}
\end{algorithm}

\subsection{Inference}
To accelerate the processing of the models with sparse activation maps, we implemented custom modifications to the XNNPACK \cite{xnnpack} inference engine.
We used TensorFlow lite (TFLite) \cite{TFLITE} built from source with XNNPACK \cite{xnnpack} as a delegate. 
Given a TFLite model, a binary mask, and layerwise sparsity levels as inputs, our inference engine computes the convolution of sparse activations. Our modifications are specific to convolutional layers only.
The full pipeline consists of three main stages: (i) custom \texttt{im2col} reshaping, (ii) dense GEMM, and (iii) custom post-processing of the dense GEMM output. 

The first step consists of reshaping the tensors into a $2$-D matrix for activations, as shown in Fig.~\ref{fig:conv_output_masking}. 
Considering that the XNNPACK \cite{xnnpack} \texttt{im2col} routine is based on an indirection buffer~\cite{dukhan2019indirect}, we developed a custom transformation to facilitate the skipping of rows of an indirection matrix.  
After this is done, the compute range of the GEMM is downsized to $output\_size~-~(sparsity~*~output\_size)$  to enable a low-rank GEMM in the following step.
In the second stage, standard GEMM is employed, utilizing a low-rank matrix of activations. However, the subsequent layer assumes dense activation, necessitating an efficient post-processing stage.  In this implementation, zeroed elements are inserted into the GEMM output based on the binary masks used in the initial stage. These modifications follow a consistent pattern across different inference engines, all designed to work with commonly used general-purpose processors. For more detailed information on the runtime modifications, please refer to Appendix A.

\section{Results}
\label{sec:results}

\begin{table}
\begin{center}
\small
\begin{tabular}{cc|ccc}
\toprule
&
\multicolumn{1}{c}{\textbf{Sparsity}} &
\multicolumn{1}{c}{\textbf{ResNet18}} & \multicolumn{1}{c}{\textbf{ResNet50}} &
\multicolumn{1}{c}{\textbf{MobileNetV2}}\\
\cline{2-5}
\parbox[t]{1mm}{\multirow{4}{*}{\rotatebox[origin=c]{90}{{\bf \footnotesize Flowers102}}}} 
& 0\% & 92.02 & 92.50 & 92.57 \\ 
& 10\% & 91.20 (-0.80) & 91.80 (-0.70) & 91.46 (-1.11) \\
& 20\% & 90.25 (-1.75) & 91.02 (-1.48) & 90.11 (-2.46) \\
& 30\% & 88.89 (-3.22) & 90.13 (-2.37) & 88.52 (-4.05) \\
\cline{2-5}
\parbox[t]{1mm}{\multirow{4}{*}{\rotatebox[origin=c]{90}{{\bf \footnotesize Food101}}}} & 0\% & 82.20 & 86.17 & 77.20 \\ 
{} & 10\% & 81.07 (-1.13) & 85.10 (-1.07) & 82.35 (-1.77) \\
{} & 10\% & 80.27 (-1.93) & 84.10 (-2.07) & 81.04 (-1.32) \\
{} & 30\% & 78.59 (-3.61) & 82.40 (-3.77) & 79.32 (-4.80) \\
\cline{2-5} 
\parbox[t]{1mm}{\multirow{4}{*}{\rotatebox[origin=c]{90}{{\bf \footnotesize CIFAR100}}}} & 0\% & 77.20 & 78.00  & 73.10 \\ 
{} & 30\% & 76.37 (-0.83) & 77.26 (-0.74) & 71.30 (-1.80) \\
{} & 30\% & 75.30 (-1.90) & 75.80 (-2.20) & 70.57 (-2.53) \\
{} & 30\% & 74.11 (-3.09) & 74.78 (-3.22) & 68.60 (-4.50) \\
\bottomrule
\end{tabular}
\end{center}
\caption{Top-1 accuracy result ($\%$) for different architectures on Flowers102, Food101, and CIFAR100 datasets. The relative inference speedups are reported in Fig.~\ref{fig:sparsity_speedup}.}
\label{tab:ic_01}
\end{table}

\begin{table}
\begin{center}
\small
\begin{tabular}{c|c|c}
\toprule
\multicolumn{1}{c}{\textbf{Sparsity}} & \multicolumn{1}{c}{\textbf{ResNet18}} & \multicolumn{1}{c}{\textbf{MobileNetV2}} \\
\hline\hline 
0\% & 70.53 & 72.19 \\ 
10\% & 70.48 (-0.05) & 70.43 (-1.76)\\
20\% & 69.42 (-1.11) & 69.94 (-2.25)\\
30\% & 67.88 (-2.65) & 67.92 (-4.27)\\
\hline
\end{tabular}
\end{center}
\caption{Top-1 accuracy results for different architectures on Imagenet dataset.  The relative inference speedups are reported in Fig.~\ref{fig:sparsity_speedup}.}
\label{tab:ic_02}
\end{table}

\begin{table}
\begin{center}
\small
\begin{tabular}{c|c|c}
\hline
\multicolumn{1}{c}{\textbf{Sparsity}} & \multicolumn{1}{c}{\textbf{VOC}} & \multicolumn{1}{c}{\textbf{Global Wheat}} \\
\hline\hline
0\% & 80.20 & 96.38 \\ \hline
10\% & 78.08 (-2.12) & 96.00 (-0.38)\\
20\% & 76.63 (-3.57) & 95.49 (-0.89)\\
30\% & 74.13 (-6.07) & 94.80 (-1.58)\\
\hline
\end{tabular}
\end{center}
\caption{mAP$_{50}$ results for YOLOv5n on VOC and Globat Wheat datasets. The relative inference speedups are reported in Fig.~\ref{fig:sparsity_speedup}.}
\label{tab:od}
\end{table}

\subsection{Training Setup}~\label{train_setup}
The proposed pipeline was validated on several image classification and object detection datasets, including CIFAR100, Flowers102, Food101, and ImageNet for classification and PASCAL VOC and Global Wheat for object detection (further details in Appendix B). We have performed experiments on ResNet18, ResNet50, and MobileNetV2 architectures for the image classification task, and used YOLOv5n~\cite{ultralytics} as a base architecture for the object detection experiments. Note that a few of the base architectures we used (e.g., MobileNetV2, YOLOv5n) were initially designed as lightweight efficient architectures, which makes it more challenging to obtain competitive latency speedup with low accuracy degradation.

For image classification, we used the training code provided by Ultralytics~\cite{ultralytics} with default values of hyperparameters except for the number of epochs (Adam optimizer, initial learning rate $10^{-4}$, $400$ epochs, batch size $64$). ImageNet pre-trained weights were used for model initialization for both the dense baseline as well as for sparse training. We set the dense training stage to stop at $10\%$ of the training steps and the freezing stage to start at $90\%$ of the steps.
For object detection experiments, the training code provided by Ultralytics~\cite{ultralytics} was also used with default values of hyperparameters. COCO pre-trained weights were used to initialize the models both for the dense baseline as well as for sparse training.

\subsection{Sparse Model Deployment}
The latency speedup from using semi-structured activation sparsity was measured on a Raspberry Pi 4B~\cite{RaspberryPi} device, featuring a quad-core ARM Cortex-A72 processor operating at $1.5$GHz, with $4$GB of RAM. We ran Ubuntu 18.04 $64$-bit OS on this platform and GNU gcc version $11.0$ for compilation. 
For deployment, we used TFLite~\cite{TFLITE} inference engine built with XNNPACK~\cite{xnnpack} delegate with custom modifications for sparse inference. 

\subsection{Sparse vs. Dense Model Performance}
In this section, we evaluate the efficacy of the semi-structured activation sparsity approach for enhancing DNN speed, prioritizing high-speed improvements at the expense of marginal accuracy degradation.
\begin{figure*}[ht]
    \centering
    \includegraphics[width=0.98\linewidth]{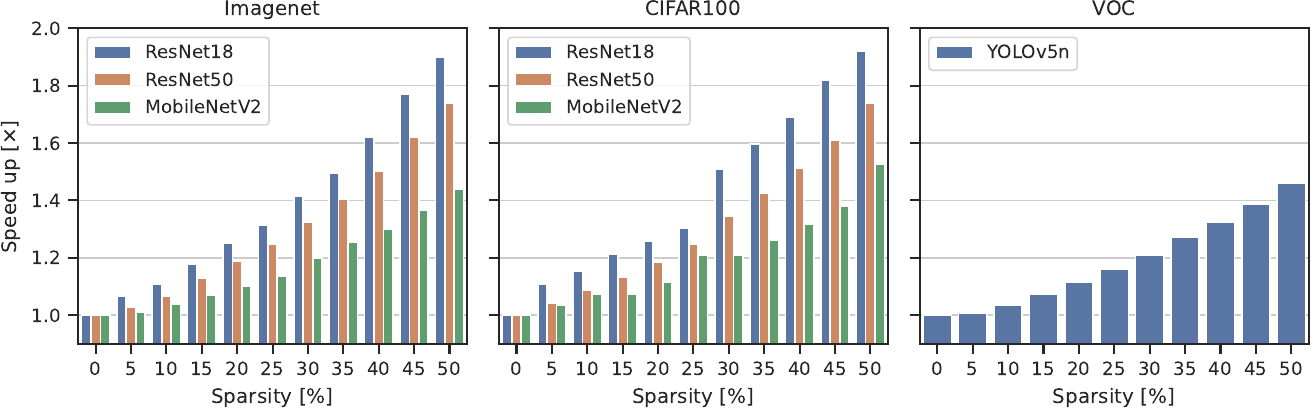}
    \caption{Speed-up vs. sparsity rate for ImageNet, CIFAR100, and VOC datasets on different architectures. Flowers102 and Food101 speed-up results are equal to those of ImageNet.}
    \label{fig:sparsity_speedup}
\end{figure*}

\subsubsection{Low Accuracy Loss Regime}
Using the same sparse training procedure, we induced the activation sparsity at three different levels $S~=~\{ 10\%, 20\%, 30\% \}$. Table~\ref{tab:ic_01} shows that the accuracy loss is low (under $2.5\%$) for the first two sparsity rate levels in image classification tasks, while it is close to $3\%$ for the highest sparsity rate chosen ($30\%$) depending on the architecture. ResNet models are found to be more resilient to activation sparsity compared to MobileNetV2, in fact, they have an average $1.82\%$ of accuracy loss instead of $2.72\%$ for MobileNetV2. On the more challenging ImageNet dataset (Table~\ref{tab:ic_02}), ResNet18 at $10\%$ sparsity rate provides almost the same accuracy ($-0.05\%$) as the dense counterpart. For clarity, we included further details on the training procedure in Appendix B.
To evaluate the generalization capabilities of our proposed compression pipeline, we carried out experiments for the object detection task using the YOLOv5n model. The obtained results on VOC and Global Wheat datasets are summarized in Table~\ref{tab:od}, showcasing the impact of compression on accuracy. Notably, results for object detection appear to be comparable to those of image classification, with limited mAP$_{50}$ degradation on a simpler dataset (Global Wheat) and higher accuracy loss observed on a more large-scale task (VOC). These findings highlight the effectiveness of our compression techniques in preserving model accuracy across different tasks.  

\subsubsection{High Speedup Regime}
In our findings, we observe a consistent trend where activation sparsity contributes to notable and reliable speed improvements throughout the network layers, with the magnitude of the speedup roughly proportional 
to the degree of activation sparsity achieved. To visually depict and quantify these results, we present Fig.~\ref{fig:sparsity_speedup}, which illustrates the end-to-end speedup outcomes for four distinct models: ResNet18, ResNet50, MobileNetV2, and YOLOv5n.

ResNet18 exhibits a nearly linear relationship between the sparsity percentage and the  speedup for all the sparsity levels. For, ResNet50, MobileNetV2, and YOLOv5n, due to the larger amount of layers and complexity, experience a slightly diminished speedup when compared to ResNet18. This slight reduction in speedup can be attributed to the presence of additional steps that involve custom \texttt{im2col} and post-processing transformations, which offset the gains obtained from reduced GEMM computations. For ResNet50, the speedup achieved is approximately $1.75\times$, while MobileNetV2 and YOLOv5n attain speedups of around $1.44\times$ and $1.46\times$, respectively, all based on $50\%$ sparsity.

In summary, our findings indicate that activation sparsity within the network layers leads to consistent and significant improvements in inference latency. The overall trend suggests that activation sparsity offers a valuable approach to enhancing the efficiency of deep learning models across a variety of architectures.

\begin{figure}[ht]
    \centering
    \includegraphics[width=0.95\linewidth]{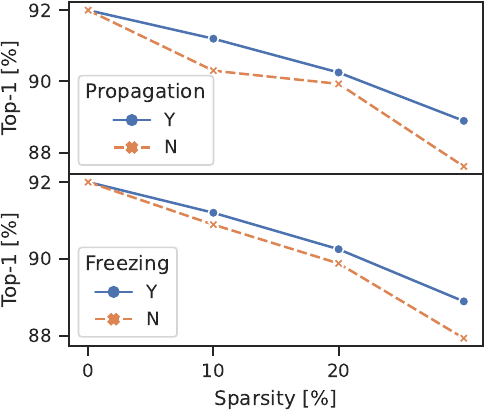}
    \caption{Ablation results for mask propagation and mask freezing for ResNet18 on Flowers102 dataset.}
    \label{fig:ablation}
\end{figure}
\subsection{Ablation Study}
To comprehensively evaluate the efficacy of our proposed sparse training scheme, we conducted two ablation studies focusing on the custom features involved to reduce accuracy loss: mask propagation and mask freezing. For both studies, we trained ResNet18 on the Flowers102 dataset using the same hyperparameters described in the Subsection~\ref{train_setup}.

\paragraph{Mask Propagation}
Figure~\ref{fig:ablation} depicts the comparison of accuracy and sparsity achieved by the ResNet18 model with and without mask propagation. The plot clearly demonstrates the advantages of employing the mask propagation method, revealing a significant improvement in the model's resilience to sparsification. The use of mask propagation provides up to $1.28\%$ of accuracy boost at $30\%$ sparsity rate and an average of $0.83\%$ for the three tested sparsity levels.
\paragraph{Mask Freezing}
The mask freezing approach ensures that the binary masks used for sparsity remain fixed during the last training epochs, thereby allowing the model to recover from accuracy loss more effectively with precise updates. This mechanism, widely used in literature~\cite{prune_or_not_to_prune}, is crucial for our training scheme where the masks are randomly changed after each step.
Figure~\ref{fig:ablation} shows the clear advantage of integrating the mask freezing method into the training process: the model trained with mask freezing showcases up to $0.96\%$ higher accuracy than the one without.

\begin{figure*}[ht]
    \centering
    \includegraphics[width=0.98\linewidth]{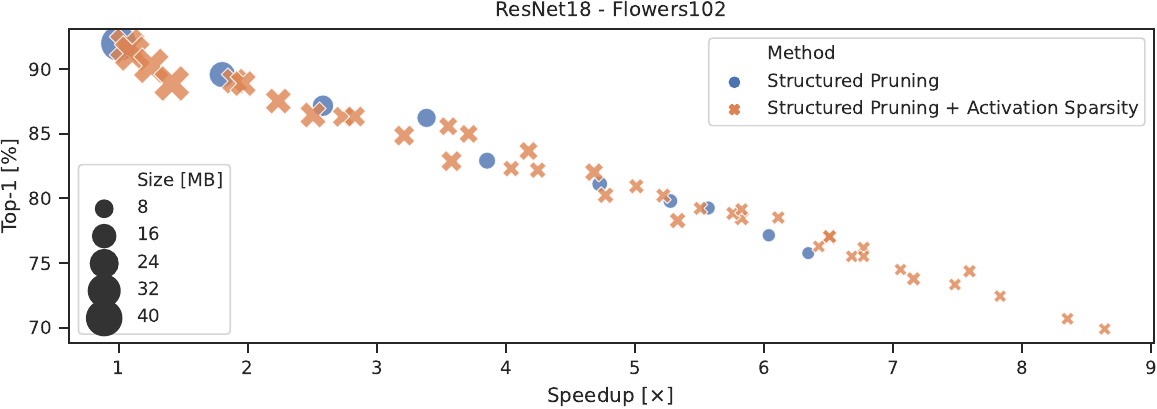}
    \caption{Latency-accuracy trade-off distribution for structured weight pruning with and without activation sparsity (ResNet18, Flowers102). A detailed table with all the numerical values is available in Appendix B.}
    \label{fig:resnet18_flowers102_pareto}
\end{figure*}
\subsection{Weight Pruning vs. Activation Sparsity}
In this section, we conduct a comprehensive comparison of our activation sparsity method with a state-of-the-art structured weight pruning technique represented by DepGraph~\cite{depgraph}. By utilizing DepGraph as a robust baseline, we aim to thoroughly assess the effectiveness and potential of our activation sparsity approach in comparison to leading compression techniques.
While the work by Kurtz et al.~\cite{inducing_activation_sparsity} appears conceptually aligned with our approach, we refrain from direct comparison due to the need for a custom sparse kernel to achieve the desired latency boost. Moreover, their research primarily focuses on higher-performance platforms, such as AWS C5.12xlarge CPU and NVIDIA K80 GPUs, rather than exploring embedded CPUs, limiting the scope of direct comparison with our solution.

Since structured weight pruning and activation sparsity can be applied independently, we decided to apply activation sparsity on models pruned using DepGraph to see the impact on performance. 
Figure~\ref{fig:resnet18_flowers102_pareto} depicts the latency vs. accuracy trade-off achievable by structured pruning with and without our proposed activation sparsity technique. We performed these experiments on ResNet18 with the Flowers102 dataset. The pruned models were obtained using the original codebase provided by DepGraph authors with different values of the speedup proxy parameter (MACs count ratio) from $2.0\times$ to $10.0\times$~\cite{depgraph}. 
Then, we induced activation sparsity in the pruned models for four different sparsity levels ($5\%$, $10\%$, $20\%$, $30\%$), using the Ultralytics training code for image classification~\cite{ultralytics}. The same training code was also used to further finetune the pruned models (without sparsity) for fair comparison.
The experimental results show that while the solely structured pruning is Pareto optimal for lower speedup rates, a combination of both techniques becomes more favorable for beyond $3.5\times$ speedup. Furthermore, while structured pruning offers high scaling ability, activation sparsity acts as a fine-grained control knob in the accuracy vs. latency solution space.
Latency measurement experiments carried out on the Raspberry Pi 4B~\cite{RaspberryPi} showcase a significant difference between the real and theoretical speedup of pruned models. A detailed table with all the different speedups is available in Appendix B.

Activation sparsity applied to pruned models shows notable performance improvements, especially for high pruning ratios. This behavior can be attributed to the understanding that models pruned beyond a certain limit may experience reduced capacity and subsequently degraded performance. In such cases, activation sparsity proves to be an effective approach by capitalizing on zeros in the activation maps, which remain independent of the model's capacity, leading to optimal results.

\section{Conclusion}
\label{sec:conclusion}

This paper presents an efficient DNN compression pipeline leveraging semi-structured activation sparsity to reduce inference latency. 
The proposed training procedure induces activation sparsity through the propagation and freezing of random spatial masks, being cognizant of element positions during GEMM-based convolutions. 
Additionally, we provide an illustrative example of a practical runtime modification integrated into XNNPACK to measure latency speedup on a Raspberry Pi 4B device.
Our experimental results showcase the impact of activation sparsity on accuracy and speedup across diverse test cases encompassing image classification and object detection tasks.
Furthermore, we demonstrate the potential to combine our compression pipeline with other structured pruning algorithms, offering enhanced accuracy-speed trade-offs, especially for high compression ratios. 
In future work, we plan to explore advanced regularization techniques to determine optimal sparsity levels across layers.

{\small
\bibliographystyle{ieee_fullname}
\bibliography{11_references}

\begin{thebibliography}{10}\itemsep=-1pt

\bibitem{ahn2020ordering}
Byung~Hoon Ahn, Jinwon Lee, Jamie~Menjay Lin, Hsin-Pai Cheng, Jilei Hou, and
  Hadi Esmaeilzadeh.
\newblock Ordering chaos: Memory-aware scheduling of irregularly wired neural
  networks for edge devices.
\newblock {\em Proceedings of Machine Learning and Systems}, 2:44--57, 2020.

\bibitem{food101}
Lukas Bossard, Matthieu Guillaumin, and Luc Van~Gool.
\newblock Food-101--mining discriminative components with random forests.
\newblock In {\em Computer Vision--ECCV 2014: 13th European Conference, Zurich,
  Switzerland, September 6-12, 2014, Proceedings, Part VI 13}, pages 446--461.
  Springer, 2014.

\bibitem{tvm}
Tianqi Chen, Thierry Moreau, Ziheng Jiang, Haichen Shen, Eddie~Q Yan, Leyuan
  Wang, Yuwei Hu, Luis Ceze, Carlos Guestrin, and Arvind Krishnamurthy.
\newblock Tvm: end-to-end optimization stack for deep learning.
\newblock {\em arXiv preprint arXiv:1802.04799}, 11(20), 2018.

\bibitem{global_wheat}
Etienne David, Simon Madec, Pouria Sadeghi-Tehran, Helge Aasen, Bangyou Zheng,
  Shouyang Liu, Norbert Kirchgessner, Goro Ishikawa, Koichi Nagasawa,
  Minhajul~A Badhon, et~al.
\newblock Global wheat head detection (gwhd) dataset: a large and diverse
  dataset of high-resolution rgb-labelled images to develop and benchmark wheat
  head detection methods.
\newblock {\em Plant Phenomics}, 2020.

\bibitem{imagenet}
Jia Deng, Wei Dong, Richard Socher, Li-Jia Li, Kai Li, and Li Fei-Fei.
\newblock Imagenet: A large-scale hierarchical image database.
\newblock In {\em 2009 IEEE conference on computer vision and pattern
  recognition}, pages 248--255. Ieee, 2009.

\bibitem{survey2020modelcompression}
Lei Deng, Guoqi Li, Song Han, Luping Shi, and Yuan Xie.
\newblock Model compression and hardware acceleration for neural networks: A
  comprehensive survey.
\newblock {\em Proceedings of the IEEE}, 108(4):485--532, 2020.

\bibitem{dong2017learning}
Xin Dong, Shangyu Chen, and Sinno Pan.
\newblock Learning to prune deep neural networks via layer-wise optimal brain
  surgeon.
\newblock {\em Advances in neural information processing systems}, 30, 2017.

\bibitem{more_is_less_gemm_dong2017more}
Xuanyi Dong, Junshi Huang, Yi Yang, and Shuicheng Yan.
\newblock More is less: A more complicated network with less inference
  complexity.
\newblock In {\em Proceedings of the IEEE conference on computer vision and
  pattern recognition}, pages 5840--5848, 2017.

\bibitem{dukhan2019indirect}
Marat Dukhan.
\newblock The indirect convolution algorithm.
\newblock {\em arXiv preprint arXiv:1907.02129}, 2019.

\bibitem{esser2019learned}
Steven~K Esser, Jeffrey~L McKinstry, Deepika Bablani, Rathinakumar Appuswamy,
  and Dharmendra~S Modha.
\newblock Learned step size quantization.
\newblock In {\em International Conference on Learning Representations}, 2019.

\bibitem{pascal_voc}
M. Everingham, S.~M.~A. Eslami, L. Van~Gool, C.~K.~I. Williams, J. Winn, and A.
  Zisserman.
\newblock The pascal visual object classes challenge: A retrospective.
\newblock {\em International Journal of Computer Vision}, 111(1):98--136, Jan.
  2015.

\bibitem{depgraph}
Gongfan Fang, Xinyin Ma, Mingli Song, Michael~Bi Mi, and Xinchao Wang.
\newblock Depgraph: Towards any structural pruning.
\newblock In {\em Proceedings of the IEEE/CVF Conference on Computer Vision and
  Pattern Recognition}, pages 16091--16101, 2023.

\bibitem{activation_map_compression_georgiadis2019}
Georgios Georgiadis.
\newblock Accelerating convolutional neural networks via activation map
  compression.
\newblock In {\em Proceedings of the IEEE/CVF Conference on Computer Vision and
  Pattern Recognition}, pages 7085--7095, 2019.

\bibitem{dropblock}
Golnaz Ghiasi, Tsung-Yi Lin, and Quoc~V Le.
\newblock Dropblock: A regularization method for convolutional networks.
\newblock {\em Advances in neural information processing systems}, 31, 2018.

\bibitem{RaspberryPi}
Google.
\newblock Raspberry pi.
\newblock \url{https://www.raspberrypi.com/products/raspberry-pi-4-model-b/},
  2023.

\bibitem{TFLITE}
Google.
\newblock Tflite.
\newblock
  \url{https://github.com/tensorflow/tensorflow/tree/master/tensorflow/lite},
  2023.

\bibitem{xnnpack}
Google.
\newblock Xnnpack.
\newblock \url{https://github.com/google/XNNPACK}, 2023.

\bibitem{skip_conv}
Amirhossein Habibian, Davide Abati, Taco~S Cohen, and Babak~Ehteshami Bejnordi.
\newblock Skip-convolutions for efficient video processing.
\newblock In {\em Proceedings of the IEEE/CVF Conference on Computer Vision and
  Pattern Recognition}, pages 2695--2704, 2021.

\bibitem{efficient_inf_engine}
Song Han, Xingyu Liu, Huizi Mao, Jing Pu, Ardavan Pedram, Mark~A Horowitz, and
  William~J Dally.
\newblock Eie: Efficient inference engine on compressed deep neural network.
\newblock {\em ACM SIGARCH Computer Architecture News}, 44(3):243--254, 2016.

\bibitem{han2015deep}
Song Han, Huizi Mao, and William~J Dally.
\newblock Deep compression: Compressing deep neural networks with pruning,
  trained quantization and huffman coding.
\newblock {\em arXiv preprint arXiv:1510.00149}, 2015.

\bibitem{han2015learning}
Song Han, Jeff Pool, John Tran, and William Dally.
\newblock Learning both weights and connections for efficient neural network.
\newblock {\em Advances in neural information processing systems}, 28, 2015.

\bibitem{distilling_hinton_2015}
Geoffrey Hinton, Oriol Vinyals, and Jeff Dean.
\newblock Distilling the knowledge in a neural network.
\newblock {\em arXiv preprint arXiv:1503.02531}, 2015.

\bibitem{hoefler2021sparsity}
Torsten Hoefler, Dan Alistarh, Tal Ben-Nun, Nikoli Dryden, and Alexandra Peste.
\newblock Sparsity in deep learning: Pruning and growth for efficient inference
  and training in neural networks.
\newblock {\em The Journal of Machine Learning Research}, 22(1):10882--11005,
  2021.

\bibitem{howard2019searching}
Andrew Howard, Mark Sandler, Grace Chu, Liang-Chieh Chen, Bo Chen, Mingxing
  Tan, Weijun Wang, Yukun Zhu, Ruoming Pang, Vijay Vasudevan, et~al.
\newblock Searching for mobilenetv3.
\newblock In {\em Proceedings of the IEEE/CVF international conference on
  computer vision}, pages 1314--1324, 2019.

\bibitem{mobilenet}
Andrew~G Howard, Menglong Zhu, Bo Chen, Dmitry Kalenichenko, Weijun Wang,
  Tobias Weyand, Marco Andreetto, and Hartwig Adam.
\newblock Mobilenets: Efficient convolutional neural networks for mobile vision
  applications.
\newblock {\em arXiv preprint arXiv:1704.04861}, 2017.

\bibitem{squeezenet}
Forrest~N Iandola, Song Han, Matthew~W Moskewicz, Khalid Ashraf, William~J
  Dally, and Kurt Keutzer.
\newblock Squeezenet: Alexnet-level accuracy with 50x fewer parameters and< 0.5
  mb model size.
\newblock {\em arXiv preprint arXiv:1602.07360}, 2016.

\bibitem{deepsparse}
Eugenia Iofinova, Alexandra Peste, Mark Kurtz, and Dan Alistarh.
\newblock How well do sparse imagenet models transfer?
\newblock {\em CoRR}, abs/2111.13445, 2021.

\bibitem{jacob2018quantization}
Benoit Jacob, Skirmantas Kligys, Bo Chen, Menglong Zhu, Matthew Tang, Andrew
  Howard, Hartwig Adam, and Dmitry Kalenichenko.
\newblock Quantization and training of neural networks for efficient
  integer-arithmetic-only inference.
\newblock In {\em Proceedings of the IEEE conference on computer vision and
  pattern recognition}, pages 2704--2713, 2018.

\bibitem{ultralytics}
Glenn Jocher, Ayush Chaurasia, and Jing Qiu.
\newblock {YOLO by Ultralytics}, Jan. 2023.

\bibitem{koubaa2023gpt}
Anis Koubaa.
\newblock Gpt-4 vs. gpt-3.5: A concise showdown.
\newblock 2023.

\bibitem{cifar100}
Alex Krizhevsky, Geoffrey Hinton, et~al.
\newblock Learning multiple layers of features from tiny images.
\newblock 2009.

\bibitem{krizhevsky2012imagenet}
Alex Krizhevsky, Ilya Sutskever, and Geoffrey~E Hinton.
\newblock Imagenet classification with deep convolutional neural networks.
\newblock {\em Advances in neural information processing systems}, 25, 2012.

\bibitem{inducing_activation_sparsity}
Mark Kurtz, Justin Kopinsky, Rati Gelashvili, Alexander Matveev, John Carr,
  Michael Goin, William Leiserson, Sage Moore, Nir Shavit, and Dan Alistarh.
\newblock Inducing and exploiting activation sparsity for fast inference on
  deep neural networks.
\newblock In {\em International Conference on Machine Learning}, pages
  5533--5543. PMLR, 2020.

\bibitem{pruning_efficient_convnet}
Hao Li, Asim Kadav, Igor Durdanovic, Hanan Samet, and Hans~Peter Graf.
\newblock Pruning filters for efficient convnets.
\newblock {\em arXiv preprint arXiv:1608.08710}, 2016.

\bibitem{pruning_efficient_convnet_li2016}
Hao Li, Asim Kadav, Igor Durdanovic, Hanan Samet, and Hans~Peter Graf.
\newblock Pruning filters for efficient convnets.
\newblock {\em arXiv preprint arXiv:1608.08710}, 2016.

\bibitem{li2022lazy}
Zonglin Li, Chong You, Srinadh Bhojanapalli, Daliang Li, Ankit~Singh Rawat,
  Sashank~J Reddi, Ke Ye, Felix Chern, Felix Yu, Ruiqi Guo, et~al.
\newblock The lazy neuron phenomenon: On emergence of activation sparsity in
  transformers.
\newblock In {\em The Eleventh International Conference on Learning
  Representations}, 2022.

\bibitem{adaptvie_sparse_inference_liu2020deep}
Ming Liu, Zhilu Zhang, Liya Hou, Wangmeng Zuo, and Lei Zhang.
\newblock Deep adaptive inference networks for single image super-resolution.
\newblock In {\em Computer Vision--ECCV 2020 Workshops: Glasgow, UK, August
  23--28, 2020, Proceedings, Part IV 16}, pages 131--148. Springer, 2020.

\bibitem{tucker_decomposition_liu2022deep}
Ye Liu and Michael~K Ng.
\newblock Deep neural network compression by tucker decomposition with
  nonlinear response.
\newblock {\em Knowledge-Based Systems}, 241:108171, 2022.

\bibitem{cnn_network_slimming}
Zhuang Liu, Jianguo Li, Zhiqiang Shen, Gao Huang, Shoumeng Yan, and Changshui
  Zhang.
\newblock Learning efficient convolutional networks through network slimming.
\newblock In {\em Proceedings of the IEEE international conference on computer
  vision}, pages 2736--2744, 2017.

\bibitem{filter_level_pruning}
Jian-Hao Luo, Jianxin Wu, and Weiyao Lin.
\newblock Thinet: A filter level pruning method for deep neural network
  compression.
\newblock In {\em Proceedings of the IEEE international conference on computer
  vision}, pages 5058--5066, 2017.

\bibitem{ma2018shufflenet}
Ningning Ma, Xiangyu Zhang, Hai-Tao Zheng, and Jian Sun.
\newblock Shufflenet v2: Practical guidelines for efficient cnn architecture
  design.
\newblock In {\em Proceedings of the European conference on computer vision
  (ECCV)}, pages 116--131, 2018.

\bibitem{wide_reduced_prec_nw}
Asit Mishra, Eriko Nurvitadhi, Jeffrey~J Cook, and Debbie Marr.
\newblock Wrpn: Wide reduced-precision networks.
\newblock {\em arXiv preprint arXiv:1709.01134}, 2017.

\bibitem{molchanov2017variational}
Dmitry Molchanov, Arsenii Ashukha, and Dmitry Vetrov.
\newblock Variational dropout sparsifies deep neural networks.
\newblock In {\em International Conference on Machine Learning}, pages
  2498--2507. PMLR, 2017.

\bibitem{molchanov2019importance}
Pavlo Molchanov, Arun Mallya, Stephen Tyree, Iuri Frosio, and Jan Kautz.
\newblock Importance estimation for neural network pruning.
\newblock In {\em Proceedings of the IEEE/CVF conference on computer vision and
  pattern recognition}, pages 11264--11272, 2019.

\bibitem{molchanov2016pruning}
Pavlo Molchanov, Stephen Tyree, Tero Karras, Timo Aila, and Jan Kautz.
\newblock Pruning convolutional neural networks for resource efficient
  inference.
\newblock {\em arXiv preprint arXiv:1611.06440}, 2016.

\bibitem{quantiztion_whitepaper}
Markus Nagel, Marios Fournarakis, Rana~Ali Amjad, Yelysei Bondarenko, Mart
  Van~Baalen, and Tijmen Blankevoort.
\newblock A white paper on neural network quantization.
\newblock {\em arXiv preprint arXiv:2106.08295}, 2021.

\bibitem{flowers}
Maria-Elena Nilsback and Andrew Zisserman.
\newblock Automated flower classification over a large number of classes.
\newblock In {\em 2008 Sixth Indian conference on computer vision, graphics \&
  image processing}, pages 722--729. IEEE, 2008.

\bibitem{activation_sparsity_gpu}
Chanyoung Oh, Junhyuk So, Sumin Kim, and Youngmin Yi.
\newblock Exploiting activation sparsity for fast cnn inference on mobile gpus.
\newblock {\em ACM Transactions on Embedded Computing Systems (TECS)},
  20(5s):1--25, 2021.

\bibitem{ren2018sbnet}
Mengye Ren, Andrei Pokrovsky, Bin Yang, and Raquel Urtasun.
\newblock Sbnet: Sparse blocks network for fast inference.
\newblock In {\em Proceedings of the IEEE Conference on Computer Vision and
  Pattern Recognition}, pages 8711--8720, 2018.

\bibitem{rhu2018compressing}
Minsoo Rhu, Mike O'Connor, Niladrish Chatterjee, Jeff Pool, Youngeun Kwon, and
  Stephen~W Keckler.
\newblock Compressing dma engine: Leveraging activation sparsity for training
  deep neural networks.
\newblock In {\em 2018 IEEE International Symposium on High Performance
  Computer Architecture (HPCA)}, pages 78--91. IEEE, 2018.

\bibitem{tan2021efficientnetv2}
Mingxing Tan and Quoc Le.
\newblock Efficientnetv2: Smaller models and faster training.
\newblock In {\em International conference on machine learning}, pages
  10096--10106. PMLR, 2021.

\bibitem{adaptive_pixelwise_activation_sparse}
Chen Tang, Wenyu Sun, Zhuqing Yuan, and Yongpan Liu.
\newblock Adaptive pixel-wise structured sparse network for efficient cnns.
\newblock {\em arXiv preprint arXiv:2010.11083}, 2020.

\bibitem{vaswani2017attention}
Ashish Vaswani, Noam Shazeer, Niki Parmar, Jakob Uszkoreit, Llion Jones,
  Aidan~N Gomez, {\L}ukasz Kaiser, and Illia Polosukhin.
\newblock Attention is all you need.
\newblock {\em Advances in neural information processing systems}, 30, 2017.

\bibitem{pruing_confusing_wang2023state}
Huan Wang, Can Qin, Yue Bai, and Yun Fu.
\newblock Why is the state of neural network pruning so confusing? on the
  fairness, comparison setup, and trainability in network pruning.
\newblock {\em arXiv preprint arXiv:2301.05219}, 2023.

\bibitem{act_sparse_conv_gpu_xu2019g}
Weizhi Xu, Yintai Sun, Shengyu Fan, Hui Yu, and Xin Fu.
\newblock Accelerating convolutional neural network by exploiting sparsity on
  gpus.
\newblock {\em ACM Transactions on Architecture and Code Optimization}, 2019.

\bibitem{distillation_zeng2000using}
Xinchuan Zeng and Tony~R. Martinez.
\newblock Using a neural network to approximate an ensemble of classifiers.
\newblock {\em Neural Processing Letters}, 12:225--237, 2000.

\bibitem{revisiting_dropout}
Yiren Zhao, Oluwatomisin Dada, Xitong Gao, and Robert~D Mullins.
\newblock Revisiting structured dropout.
\newblock {\em arXiv preprint arXiv:2210.02570}, 2022.

\bibitem{prune_or_not_to_prune}
Michael Zhu and Suyog Gupta.
\newblock To prune, or not to prune: exploring the efficacy of pruning for
  model compression.
\newblock {\em arXiv preprint arXiv:1710.01878}, 2017.

\end{thebibliography}
}

\ifarxiv \clearpage \section*{\LARGE Appendix}
\appendix
\label{sec:appendix}

\section{Inference Engine Modification}
In the context of runtime modifications for activation sparsity inference, we used XNNPACK~\cite{xnnpack} as our inference engine and made minor adaptations to ensure robust support for inference. Algorithm~\ref{alg:custom_inference} illustrates a simplified pseudocode representation of these crucial modifications. 
Additional information about these referenced functions is readily available within the code repository~\cite{xnnpack}.
The implementation comprises three stages: (i) a custom indirection-based \texttt{im2col}, (ii) a standard dense GEMM, and (iii) custom post-processing components. 

\begin{algorithm}[h!]
\small
\SetAlgoLined
\SetKwProg{Function}{Function}{:}{end}
\SetKwFunction{customIndirection}{customIndirection}
\SetKwFunction{postProcess}{postProcess}

\BlankLine
\Function{xnn\_indirection\_init\_conv2d\_sparse}{
    indirection\_buffer $\leftarrow$ empty list\;

    \For{out\_y \KwTo output\_height}{
        \For{out\_x \KwTo output\_width}{
            \If{mask[out\_x][out\_y] $==$ 1}{
              indirection\_buffer[index] $\leftarrow$ (const void*) ((uintptr\_t) input + base\_address + offset)\;
            }
        }
    }
    \Return indirection\_buffer\;
}
\BlankLine
\BlankLine
\BlankLine
\Function{post\_process\_conv2d\_sparse}{
  outch\_size = output\_channels ~*~ sizeof(float)\;

     \For{out\_y $\rightarrow$ output\_height - 1 \KwTo 0}{
    \For{out\_x $\rightarrow$ output\_width - 1 \KwTo 0}{
            \If{mask[out\_x][out\_y] $==$ 0}{

           memset(op$\rightarrow$ output[out\_y~*~op$\rightarrow$ output\_height~+~out\_x], 0, outch\_size)\;
            
             }
            \Else{

           memcpy(output[out\_y~*~ output\_height~+~out\_x], output[id], 
 outch\_size)\;
 id $\leftarrow$ id - 1\;
           
           }
        }
    }
  \Return output;
}
\caption{Inference Engine Modification}
\label{alg:custom_inference}
\end{algorithm}
At first, the {\small \texttt{xnn\_indirection\_init\_conv2d\_sparse}} (lines $1$-$11$) function illustrates our custom approach for efficiently skipping rows within an indirection matrix. 
Within the indirection-based \texttt{im2col} function, we deviate from memory-intensive transformations and instead, store input value pointers in the indirection buffer. This strategy adheres to the loop structure commonly employed in the standard \texttt{im2col} transformation.
Diverging from the original procedure, our implementation skips converting the entire convolution patch into a single row when mask values equate to $0$  (line $5$).
To accurately allocate the appropriate input value pointer to a designated position (\texttt{index}) within the \texttt{indirection\_buffer}, we employed the \texttt{base\_address} and \texttt{offset} variables. These variables are computed according to the \texttt{conv2d} parameters.
Upon completing this initial step, the computation range of the GEMM is reduced to $output\_size - (sparsity \times output\_size)$.
The GEMM function operates as a subroutine that efficiently conducts dense matrix multiplication between weight and activation values. This function remains unaltered, with no modifications made to the underlying kernel.

Lastly, the \texttt{post\_process\_conv2d\_sparse} function (lines $12$-$26$) manages the output for the subsequent layer, incorporating a transformation that involves the insertion of zeros based on the corresponding mask value. When the mask value is $0$ (lines $16$-$18$), the function inserts zeros. Alternatively, when the mask value is $1$ (lines $19$-$21$), data is copied from one position to another within the same output channel size, denoted as \texttt{outch\_size} in the algorithm. This customized procedure is tailored for post-processing the output subsequent to low-rank GEMM operations, and it is invoked within the \texttt{xnn\_run\_operator} method~\cite{xnnpack}.

\section{Additional Details on Experiments}

\subsection{Visualization}
\begin{figure}[h!]
    \centering
    \includegraphics[width=0.98\linewidth]{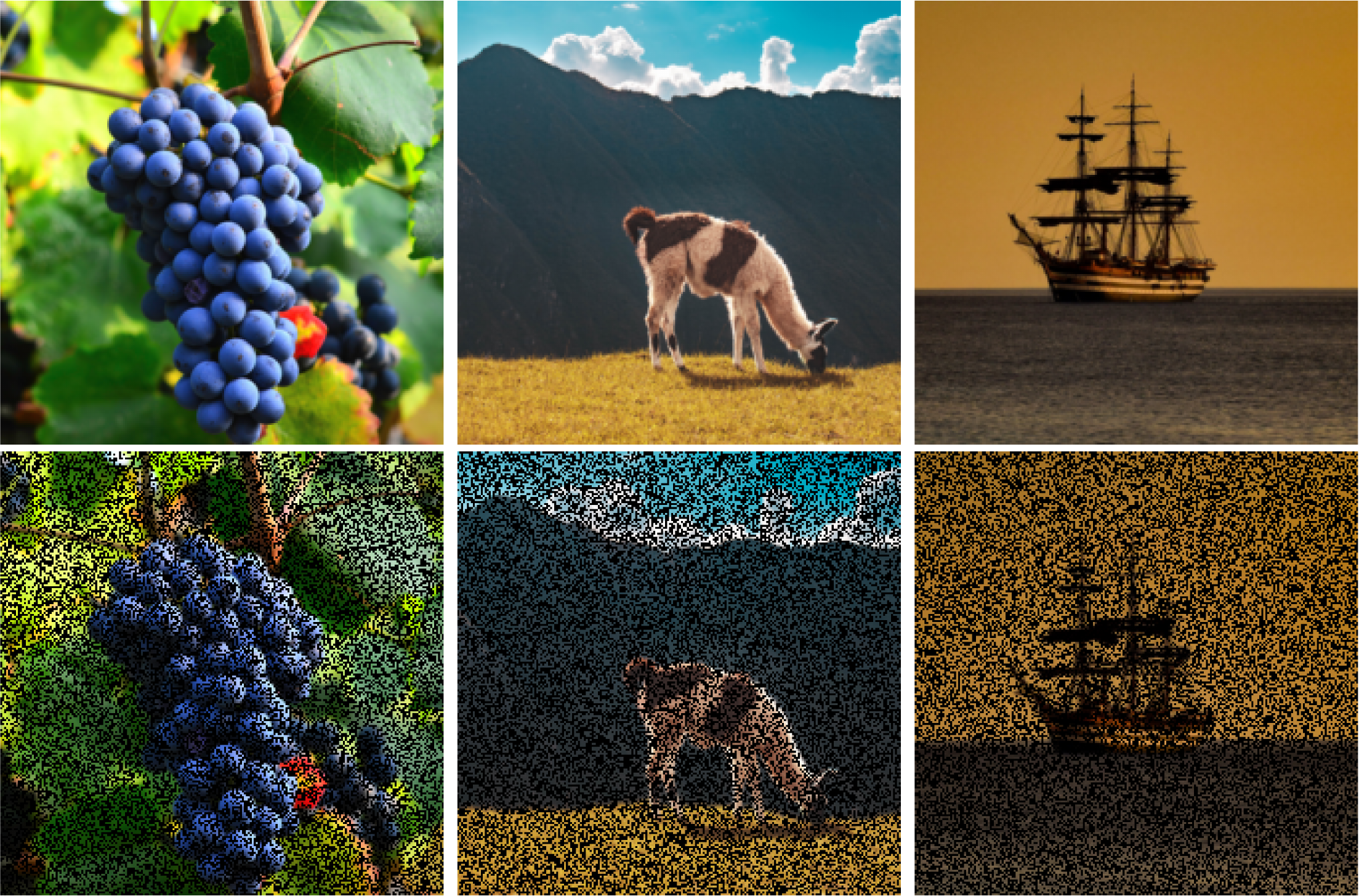}
    \caption{Even with $30\%$ sparsity induced in these three pictures, the main content remains visible and comprehensible to human eyes.}
    \label{fig:sparse_pics}
\end{figure}
In Figure~\ref{fig:sparse_pics}, we illustrate a visual comparison of three standard $224\times224$ images, both before and after the application of a $30\%$ sparsity constraint. Notably, even when subjected to a substantial level of induced sparsity, the core content remains discernible and comprehensible to the human observer. While certain finer details may be sacrificed due to the reduction in non-zero pixel values, the fundamental subject matter and distinctive characteristics of each image endure. This observation suggests that when visual content remains clear to the human eye, deep neural networks are likely to recognize the semantic content of the images as well, particularly in scenarios with lower levels of activation sparsity.

\subsection{Datasets}
{\bf CIFAR-100 ~\cite{cifar100}} :
It comprises $60,000$ RGB images, each measuring $32\times32$ pixels, and annotated with $100$ distinct labels with $45,000$ training, $5,000$ validation, and $10,000$ testing samples.

{\bf Flowers102~\cite{flowers}}:
This dataset is a collection of $102$ categories of flower species, with each category containing a variable number of RGB images. Each image is of arbitrary size and comes with appropriate labels indicating the corresponding flower species. We used $224 \times 224$ image resolution.

{\bf Food101~\cite{food101}}:
It comprises a diverse set of food images spanning $101$ distinct classes, the dataset offers a valuable resource for food recognition tasks. Each RGB image in the dataset is associated with a specific food category. We used $224 \times 224$ image resolution.

{\bf ImageNet~\cite{imagenet}}:
This dataset comprises $1$M of RGB images belonging to a vast array of classes, enabling in-depth evaluation of image classification capabilities. The pipeline leveraged subsets of the ImageNet dataset, ensuring a representative and diverse range of images for training, validation, and testing purposes.

{\bf PASCAL VOC~\cite{pascal_voc}}:
The PASCAL VOC dataset, derived from the PASCAL Visual Object Classes Challenge, encompasses $15,870$ RGB images with 37,813 object annotations for $20$ different categories. The pipeline adhered to the recommended approach outlined in, utilizing the VOC07 and VOC12 trainval data for training, while the VOC07 dataset was employed for testing purposes. We used $480 \times 480$ image resolution.

{\bf Global Wheat~\cite{global_wheat}}: The Global Wheat Head Dataset is a collection of images designed to support the development of accurate wheat head detection models for applications in wheat phenotyping and crop management. The dataset contains over $3000$ images in the training set, and approximately $1000$ images for validation taken in different regions. We train and evaluate with $480 \times 480$ image resolution in our experiments.

\subsection{Training}
For ResNet models, we did not induce activation sparsity in the pointwise downsample layers ($1\times1$ convolutional kernels), as their overall contribution to the runtime is negligible. Furthermore, this allows the model to recover a small amount of accuracy (e.g., around $0.2\%$ for ResNet18 on Flowers102 dataset).
Figures~\ref{fig:train_curve}~and~\ref{fig:train_curve_zoomed} report an example of the training curves to offer comprehensive insights into the proposed method's learning behavior, providing a deeper understanding of the training dynamics and overall training performance.
\begin{figure*}[ht!]
\centering
\includegraphics[width=0.98\linewidth]{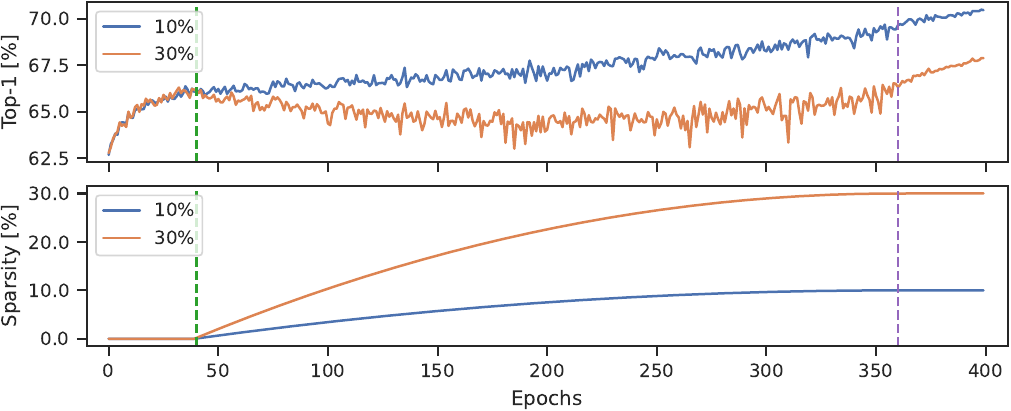}
\caption{Training curves for ResNet18 on the ImageNet dataset for two different sparsity levels. The two vertical lines split the training curve according to the three different stages. From epoch $0$ to epoch $40$ (green line) the dense pretraining steps, from epoch $40$ to epoch $360$ (purple line) the sparse training steps with variable random masking, and, at last, from epoch $360$ to the end the mask freezing stage.}
\vspace{5mm}
\label{fig:train_curve}
\includegraphics[width=0.98\linewidth]{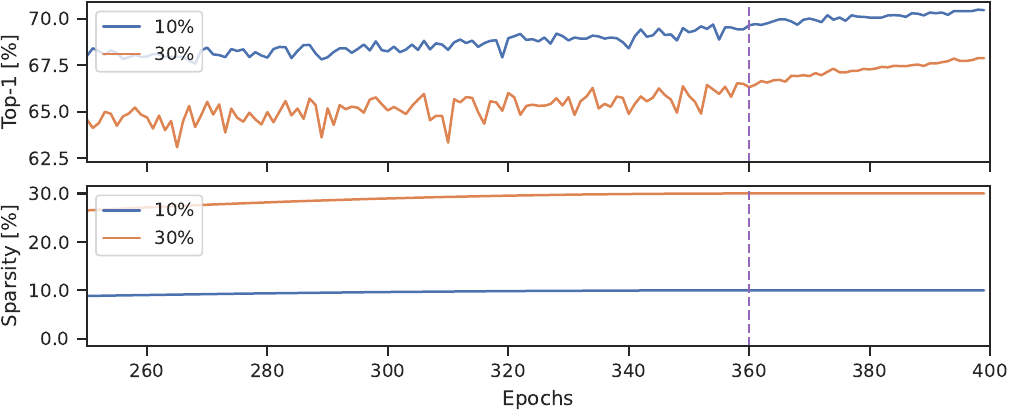}
\caption{Training curves for ResNet18 on the ImageNet dataset for two different sparsity levels. The same training curves of Fig.~\ref{fig:train_curve}, here zoomed in on the last epochs to better show the effects of the mask freezing stage (from epoch $360$ to $400$).}
\label{fig:train_curve_zoomed}
\end{figure*}

\begin{table*}[hb!]
\center
\large
\begin{tabular}{cc|cc||cc|cc|cc|cc}
\multicolumn{4}{c}{\textbf{Structured Weight Pruning}} & \multicolumn{8}{c}{\textbf{Structured Weight Pruning + Activation Sparsity}}\\
 \toprule
\multicolumn{2}{c}{\textbf{Depgraph} \cite{depgraph}} & \multicolumn{2}{c||}{\textbf{Fine-tuned} \cite{ultralytics}} & \multicolumn{2}{c} {\textbf{5\%}} & \multicolumn{2}{c} {\textbf{10\%}} & \multicolumn{2}{c} {\textbf{20\%}} & \multicolumn{2}{c} {\textbf{30\%}}\\
\hline\hline
{S} & {A} &  {S} & {A} & {OS} & {A} & {OS} & {A} & {OS} & {A} & {OS} & {A} \\
\hline
1.0 & / & 1.0 & 92.02 & 1.07 & 91.80 & 1.11 & 91.20 & 1.25 & 90.25 & 1.41 & 88.89 \\
2.0 & 89.46 & 1.8 & 89.58 & 1.90 & 89.07 & 1.96 & 88.88 & 2.24 & 87.53 & 2.51 & 86.45 \\
3.0 & 86.27 & 2.6 & 87.17 & 2.74 & 86.31 & 2.83 & 86.34 & 3.21 & 84.86 & 3.58 & 82.88 \\
4.0 & 85.18 & 3.4 & 86.23 & 3.55 & 85.58 & 3.71 & 84.99 & 4.18 & 83.67 & 4.68 & 82.03 \\
5.0 & 81.93 & 3.9 & 82.92 & 4.04 & 82.31 & 4.25 & 82.19 & 4.77 & 80.24 & 5.33 & 78.29 \\
6.0 & 79.87 & 4.7 & 81.12 & 5.01  & 80.94 & 5.22 & 80.22 & 5.83 & 78.47 & 6.51 & 77.01 \\
7.0 & 79.44 & 5.3 & 79.80 & 5.51 & 79.23 & 5.76 & 78.84 & 6.51 & 77.05 & 7.16 & 73.78 \\
8.0 & 78.27 & 5.6 & 79.26 & 5.83 & 79.15 & 6.11 & 78.52 & 6.77 & 76.17 & 7.59 & 74.37 \\
9.0 & 76.01 & 6.0 & 77.15 & 6.42 & 76.29 & 6.68 & 75.52 & 7.48 & 73.34 & 8.35 & 70.69 \\
10.0 & 74.65 & 6.3 & 75.77 & 6.68 & 75.52 & 6.77 & 74.50 & 7.26 & 72.44 & 7.59 & 69.90\\
\bottomrule
\end{tabular}
\vspace{2mm}
\caption{Latency-accuracy results for structured pruning without (first four columns) and with activation sparsity (last eight columns) for ResNet18 on the Flowers102 dataset. For each pair of structured pruning columns, we report speedup (S, $\times$) and top-1 accuracy (A, $\%$). The first group shows the results obtained using the original training code of Depgraph~\cite{depgraph} with the estimated speedups, while the second one shows the results obtained with further fine-tuning using Ultralytics training code~\cite{ultralytics} with the real speedups measured on the device.
For each pair of columns of structured pruning with activation sparsity, we report overall speedup (OS, $\times$) and top-1 accuracy (A, $\%$) at different levels of sparsity, trained using the same Ultralytics training code~\cite{ultralytics}. }
\end{table*} \fi

\end{document}